\documentclass[a4paper,11pt]{article}
\usepackage[margin=1in]{geometry}
\usepackage[centertags]{amsmath}
\usepackage{amssymb,amsmath,amsthm,amscd,latexsym}
\usepackage{amsfonts}
\usepackage[mathscr]{eucal}
\usepackage{color}
\usepackage{graphicx}
\usepackage{multirow}
 \newtheorem{theorem}{Theorem}[section]

 \theoremstyle{definition}
 
 \theoremstyle{remark}

 \numberwithin{equation}{section}
\begin{document}

\title{A Novel Genetic Search Scheme Based on Nature-Inspired Evolutionary Algorithms for Self-Dual Codes}
\author{
Adrian Korban \\
Department of Mathematical and Physical Sciences \\
University of Chester\\
Thornton Science Park, Pool Ln, Chester CH2 4NU, England \\
Serap \c{S}ahinkaya \\
Tarsus University, Faculty of Engineering \\ Department of Natural and Mathematical Sciences \\
Mersin, Turkey \\
Deniz Ustun \\
Tarsus University, Faculty of Engineering \\ Department of Computer Engineering \\
Mersin, Turkey}

\maketitle

\begin{abstract}
In this paper, a genetic algorithm, one of the evolutionary algorithms optimization methods, is used for the first time for the problem of finding extremal binary self-dual codes. We present a comparison of the computational times between a genetic algorithm and a linear search for different size search spaces and show that the genetic algorithm is capable of finding binary self-dual codes significantly faster than the linear search. Moreover, by employing a known matrix construction together with the genetic algorithm, we are able to obtain new binary self-dual codes of lengths 68 and 72 in a significantly short time. In particular,
we obtain 11 new extremal binary self-dual codes of length 68 and 17 new binary self-dual codes of length 72.
\end{abstract}


\section{Introduction}

There has been a great interest and focus on constructing binary self-dual codes of different lengths with new weight enumerators. This is partly due to the fact that self-dual codes have a strong connection to structures such as designs, lattices or invariant polynomials to name a few, please see \cite{Doughertybook2} for many interesting examples of such connections. We recall that a binary self-dual code of length $2n$ is generated by (up to equivalence) a matrix of the form $[I|A],$ where $A$ is an $n \times n$ block matrix. As stated in \cite{Kaya2}, the randomness of the matrix $A$ makes for an impractical search field of $2^{n^2}$ and even when the orthogonality relations are factored in, we still have a search field of size $2^{n(n+1)/2},$ which is still considerably large. Many researchers have utilized particular types of matrices in an effort to reduce the search field and still find binary self-dual codes with new weight enumerators. In this work, we employ a genetic algorithm (GA) when performing the searches for binary self-dual codes. This algorithm turns out, as we show in this work, to a be an extremely good tool for coping with large search fields when finding codes  that have a generator matrix of the $[I|A]$ form. By a good tool, we mean that this algorithm finds codes significantly faster when compared to a standard linear search.

Many self-dual codes with generator matrices of the $[I|A]$ form were found by using a linear search (LS) method, for example, please see \cite{Korban1, Korban2, Korban3, Gildea2, Korban4}. There is no indication in the literature that people have tried the genetic algorithm to search for extremal binary self-dual codes in the past. Although a linear search  is a powerful method for finding solutions in the small-size search space, it takes long time to obtain the optimal point when the search space grows.  On the other hand, evolutionary algorithms (EA) optimization  methods have received much attention in recent years due to their success in many problems with large parameters. The EA optimization methods have been used to find the optimum solutions of differentiable and continuous NP-hard problems for decades. These methods, named also as nature-inspired techniques, use the heuristic approaches that include different calculations for finding the optimal points. In this paper, genetic algorithm (GA), a search based on the nature-inspired evolutionary algorithm, is used for the first time to search for self-dual codes that have generator matrices of the $[I|A]$ form. We show that the heuristic approaches outperform the linear searches, by performing a search for extremal binary self-dual codes using the two approaches.

The rest of the paper is organised as follows. In Section~2, we give a brief history of the genetic algorithm and describe how it copes with large size search spaces. We also give preliminary definitions and results on self-dual codes, special matrices, group rings and the well known construction method from \cite{Gildea1}. In Section~3, we employ the construction method from Section~2 and run a comparison of the computational times between the genetic algorithm and the linear search. We show that the genetic algorithm outperforms the linear search especially when the search field grows. In Section~4, we employ the construction method from Section~2 and attempt to search for extremal binary self-dual codes of lengths 68 and 72 by using the genetic algorithm. As a result, we find 11 new extremal binary self-dual codes of length 68 and 17 new binary self-dual codes of length 72 in a very short time. Some of the codes we obtain have weight enumerators with parameters that were not known in the literature before or are very rare. We finish with concluding remarks and directions for possible future research.

\section{Preliminaries}

\subsection{Genetic Algorithm}

The evolutionary algorithms (EA) are a very effective and powerful algorithms that have been used for solving hard problems for which the solutions can not be directly found in the polynomial time, such as the classical NP-Hard problems from \cite{Goldberg1} \cite{Goldberg2} and \cite{Holland}. They also achieve good results for the problems in which solution steps take exhaustively long time. A genetic algorithm (GA), a population-based meta-heuristic optimization method, was first introduced by Holland  \cite{Holland} and has been one of the widely used optimization methods based on evolutionary algorithms. The GA uses mutation, crossover, and selection procedures that are inspired by mechanisms of the biological evaluation. In GA, the population consists of a set of the chromosomes, which is also called a candidate solutions.

 \begin{figure*}[h!]
\centering
\includegraphics[width=40mm]{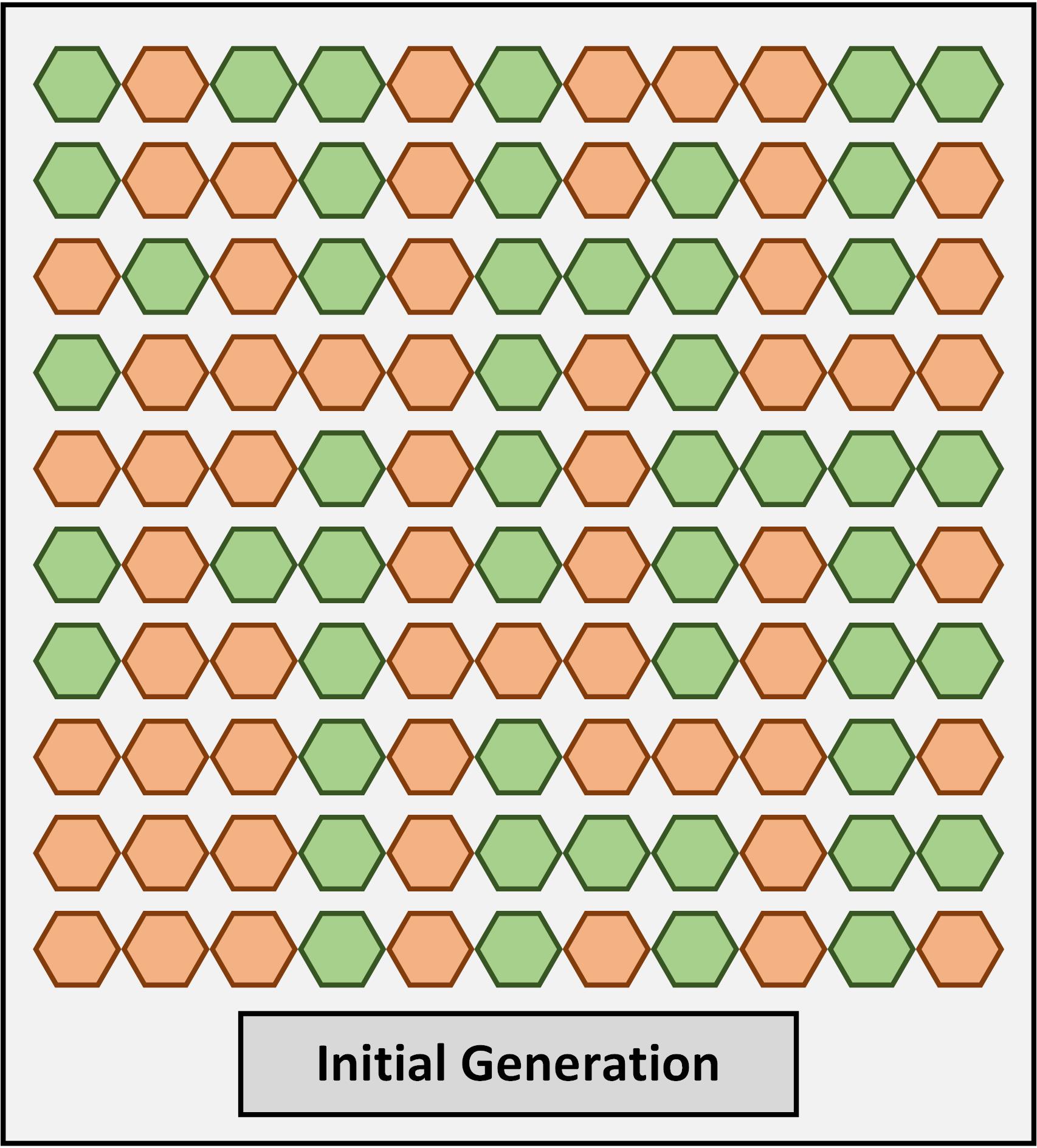}
\caption{The Initial Phase \label{Initial}}
\end{figure*}

At first, a chromosomes pool which is represented in Figure \ref{Initial}, is randomly generated for an initial population. The fitness value for each chromosome is calculated by an objective function:

\begin{equation}\label{obj}
 Obj_i=\frac{1}{d_i},
\end{equation}
here $d_i$ is the minimum distance for the $i$-th code.  A set of parents is chosen from the chromosome population for producing the new child genes via genetic operators. The fitness values of the child genes are calculated like in the initial phase. The old chromosomes  are replaced by new chromosomes according to a certain selection process based on greedy selection approach. The GA is repeated until a requested termination criteria.

\begin{figure*}[h!]
\centering
\includegraphics[width=95mm]{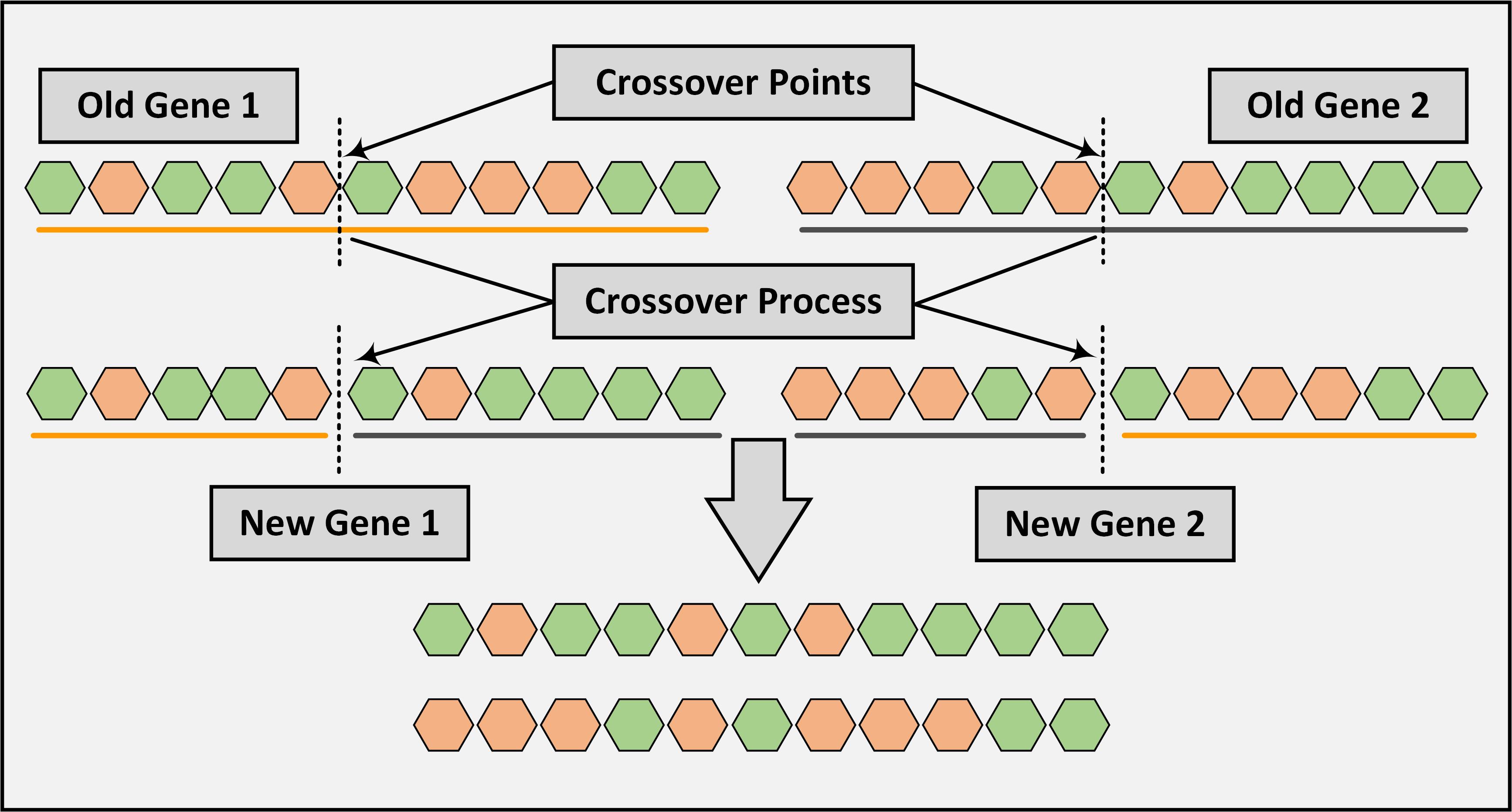}
\caption{The  Crossover Phase \label{Crossover}}
\end{figure*}

In the crossover phase, a recombination process is performed and child chromosomes are generated by using some pieces of both parents genetic material. In the crossover process, a probability term is set for determining operation rate. A number of variations such as single-point crossover, double-point crossover and uniform-crossover are used for the crossover phase of GA. A single-point crossover is the simplest form and an example is given in Figure \ref{Crossover}. Two chromosomes in a population are randomly selected by using the selection scheme given in Figure \ref{Crossover}. A point is randomly chosen on the chromosomes for the crossover process and then the parts of the two chromosomes beyond the selected point are swapped to generate a child chromosomes. The double-point crossover process is the same as the single-point crossover. However, the only difference is that two different crossover-points are randomly chosen instead of one. Single and double crossover determine cross points where chromosomes can be divided. A uniform-crossover occurs a scheme to make each bit a potential crossover point.
A binary string scheme with the same length chromosome, which crossovers process will be performed, is generated randomly. A parent chromosome supplies the child chromosome with the associated bit. In a uniform-crossover phase, the binary string scheme is examined for each bit location. If the random string contains a “1” at that location, the corresponding bits of the parent chromosomes are swaped. If the random bit is “0” no exchange takes place.

\begin{figure*}[h!]
\centering
\includegraphics[width=85mm]{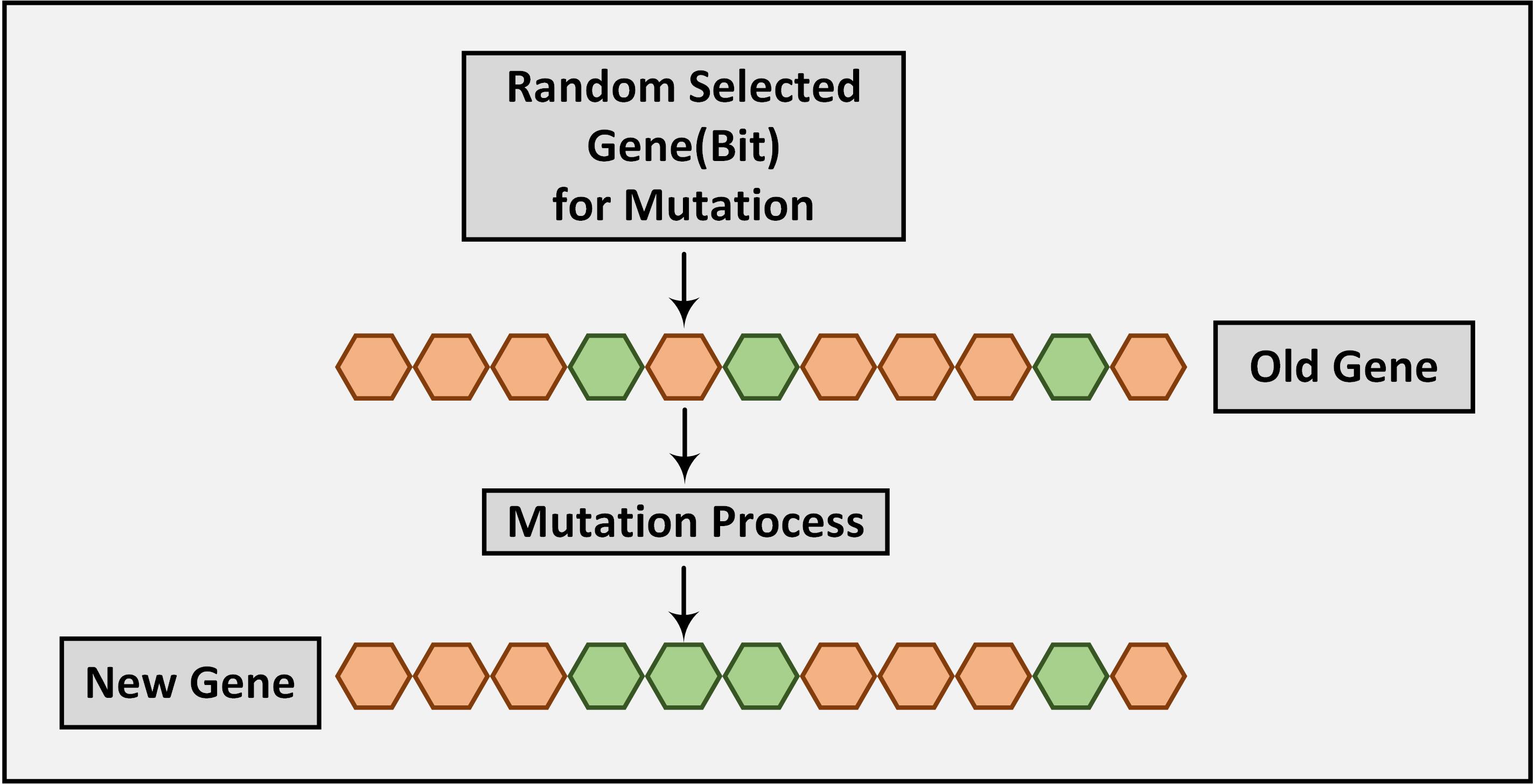}
\caption{The Mutation Phase \label{Mutation}}
\end{figure*}

In the mutation phase, the variations which provide the diversity in the gene population are introduced into the chromosome by using the mutation operator. The variations positively assist the performance of GA in global and local searching processes. The mutation process is based on a small probability \textit{p} value that initially is set, it performs from time to time. Also, the gene bit to be mutated is chosen randomly. A random number is produced and if the random number is smaller than the probability value (\textit{p}), then each bit of a bit-string on a chromosome is replaced by a randomly generated bit. An example of mutation process on the eleven bits is given in Figure \ref{Mutation}.

After generating the sub-population (child genes, offspring) in the crossover and mutation phases, the old chromosomes in the population are exchanged by  the new chromosomes from sub-population. The selection method  is an elitist strategy that one or a few of the best chromosomes   are transferred into the succeeding generation.

\begin{figure*}[h!]
\centering
\includegraphics[width=140mm]{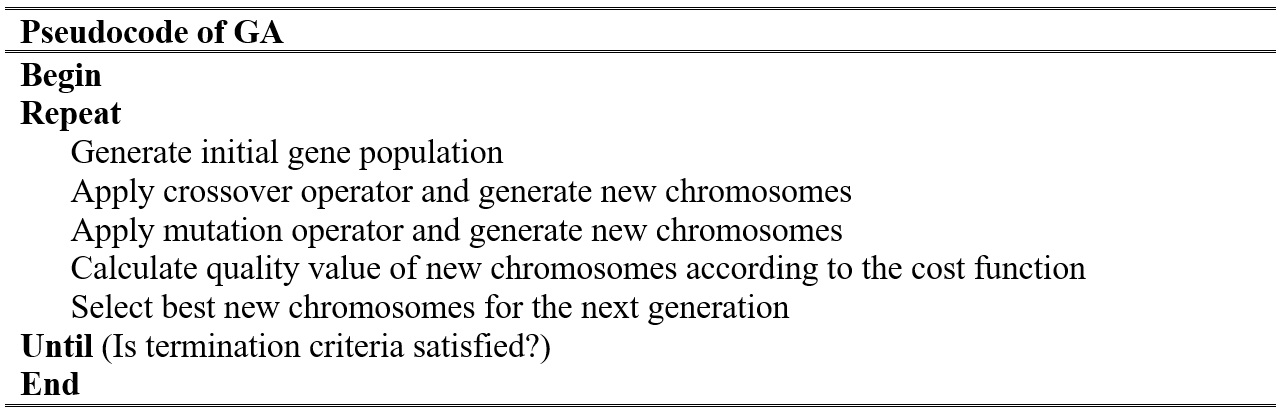}
\caption{The pseudocode of GA \label{pseudocode}}
\end{figure*}

\begin{figure*}[h!]
\centering
\includegraphics[width=140mm]{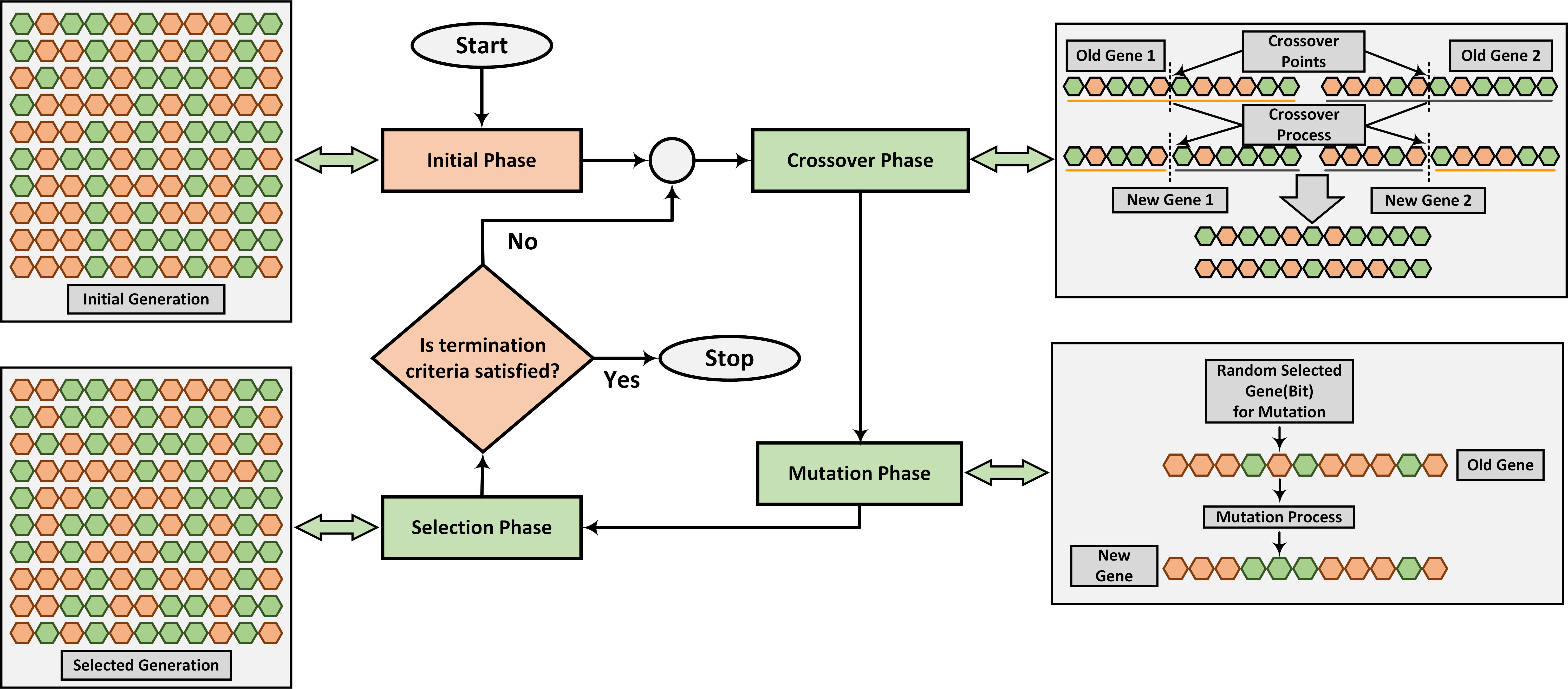}
\caption{The  Flowchart of GA algorithm \label{GAAlgorithm}}
\end{figure*}

In GA,  the above mentioned mutation, crossover and selection phases repeats iteratively until a certain condition value is obtained which  is set at the initial stage. The best chromosomes are saved for the next generation at the end of each iteration.  A flowchart  and pseudo-code of GA, giving details of operation of algorithm phases, are given in Figure \ref{GAAlgorithm} and Figure \ref{pseudocode}.

\newpage

\subsection{Binary Self-Dual Codes}

A binary code $C$ of length $n$ is said to be self-dual if $C^{\bot}=C.$ Self-dual codes are necessarily linear and self-orthogonal. Furthermore, their dimension must be $n/2,$ which means $n$ must be even. All codewords of a self-dual code have even weights. If all the weights of all codewords in the self-dual code $C$ are divisible by $4,$ then $C$ is called a Type II (or doubly even) code. Otherwise $C$ is called a Type I (singly even) code. Binary self-dual codes have bounds on their minimum distances:

\begin{theorem}
(\cite{Rains1}) Let $d_I(n)$ and $d_{II}(n)$ be the minimum distance of a Type I and Type II binary code of length $n,$ respectively. Then
\begin{equation*}
d_{II}(n) \leq 4\lfloor \frac{n}{24} \rfloor+4
\end{equation*}
and
\begin{equation*}
d_{I}(n)\leq
\begin{cases}
\begin{matrix}
4\lfloor \frac{n}{24} \rfloor+4 \ \ \ if \ n \not\equiv 22 \pmod{24} \\
4\lfloor \frac{n}{24} \rfloor+6 \ \ \ if \ n \equiv 22 \pmod{24}.%
\end{matrix}%
\end{cases}%
\end{equation*}
\end{theorem}

Self-dual codes meeting these bounds are called extremal. Finding extremal binary self-dual codes of certain lengths is a relevant problem in Coding Theory that attracts a lot of attention.

\subsection{Special Matrices, Group Rings and the Constructions}

We start this section by recalling the definitions of some special matrices
which we use later in our work. A circulant matrix is one where each row is
shifted one element to the right relative to the preceding row. We label the
circulant matrix as $A=circ(\alpha_1,\alpha_2\dots , \alpha_n),$ where $%
\alpha_i$ are the ring elements. A reverse circulant matrix is one where
each row is shifted one element to the left relative to the preceding row.
We label the reverse circulant matrix as $A=rcirc(\alpha_1,\alpha_2,\dots,%
\alpha_n).$ A block-circulant matrix is one where each row contains blocks
which are square matrices. The rows of the block matrix are defined by
shifting one block to the right relative to the preceding row. We label the
block-circulant matrix as $CIRC(A_1,A_2,\dots A_n),$ where $A_i$ are
the $k \times k$ matrices over the ring $R.$ A symmetric matrix is a square
matrix that is equal to its transpose. The transpose of a matrix $A,$
denoted by $A^T,$ is a matrix whose rows are the columns of $A.$

In our main construction we also apply matrices which come from group ring
elements, we therefore now state the necessary
definitions for group rings.

While group rings can be given for infinite rings and infinite groups, we
are only concerned with group rings where both the ring and the group are
finite. Let $G$ be a finite group of order $n$, then the group ring $RG$
consists of $\sum_{i=1}^n \alpha_i g_i$, $\alpha_i \in R$, $g_i \in G.$

Addition in the group ring is done by coordinate addition, namely
\begin{equation}
\sum_{i=1}^n \alpha_i g_i +\sum_{i=1}^n \beta_i g_i =\sum_{i=1}^n (\alpha_i
+ \beta_i)g_i.
\end{equation}
The product of two elements in a group ring is given by
\begin{equation}
\left(\sum_{i=1}^n \alpha_i g_i \right)\left(\sum_{j=1}^n \beta_j g_j
\right)= \sum_{i,j} \alpha_i \beta_j g_i g_j.
\end{equation}
It follows that the coefficient of $g_k$ in the product is $\sum_{g_i
g_j=g_k} \alpha_i \beta_j.$

The following construction of a matrix was first given for codes over fields
by Hurley in \cite{Hurley1}. It was extended to Frobenius rings in \cite{Dougherty3}. Let $%
R$ be a finite commutative Frobenius ring and let $G=\{g_1,g_2,\dots,g_n\}$
be a group of order $n$. Let $v=\alpha_{g_1}g_1+\alpha_{g_2}g_2+\dots
+\alpha_{g_n}g_n \in RG.$ Define the matrix $\sigma(v) \in M_n(R)$ to be
\begin{equation}
\sigma(v)=%
\begin{pmatrix}
\alpha_{g_1^{-1}g_1} & \alpha_{g_1^{-1}g_2} & \alpha_{g_1^{-1}g_3} & \dots &
\alpha_{g_1^{-1}g_n} \\
\alpha_{g_2^{-1}g_1} & \alpha_{g_2^{-1}g_2} & \alpha_{g_2^{-1}g_3} & \dots &
\alpha_{g_2^{-1}g_n} \\
\vdots & \vdots & \vdots & \vdots & \vdots \\
\alpha_{g_n^{-1}g_1} & \alpha_{g_n^{-1}g_2} & \alpha_{g_n^{-1}g_3} & \dots &
\alpha_{g_n^{-1}g_n}%
\end{pmatrix}%
.
\end{equation}

We note that the elements $g_1^{-1}, g_2^{-1}, \dots, g_n^{-1}$ are the
elements of the group $G$ in a some given order.

We are now ready to state the matrix construction that we employ later in this work to search for binary self-dual codes of different lengths. We stress that this is not a new construction, it was first introduced in \cite{Gildea1}.

\begin{theorem}\label{MainMatrixConstruction}
(\cite{Gildea1}) Let $G=\{g_1,g_2,g_3,\dots,g_n\}$ be a group of order $n$ and $R$ be a commutative Frobenius ring of characteristic 2. Let $v=a_1g_1+a_2g_2+\dots +a_ng_n \in RG$ be an element of the group ring $RG.$ Then the generator matrix of the following form
\begin{equation}
[I \ | \ \sigma(v)]
\end{equation}
generates a self-dual codes over $R$ if and only if $\sigma(v)\sigma(v)^T=I_n.$
\end{theorem}

The advantage of the above construction is that for different groups $G$, the map $\sigma$ produces different matrices that are fully defined by the elements that appear in the first row. In \cite{Gildea1}, the authors have considered groups of orders 8 and 16 and different alphabets to search for codes of length 64.

\section{The Comparison of Genetic Algorithms with Linear Search}

\begin{figure*}[h!]
\centering
\includegraphics[width=95mm]{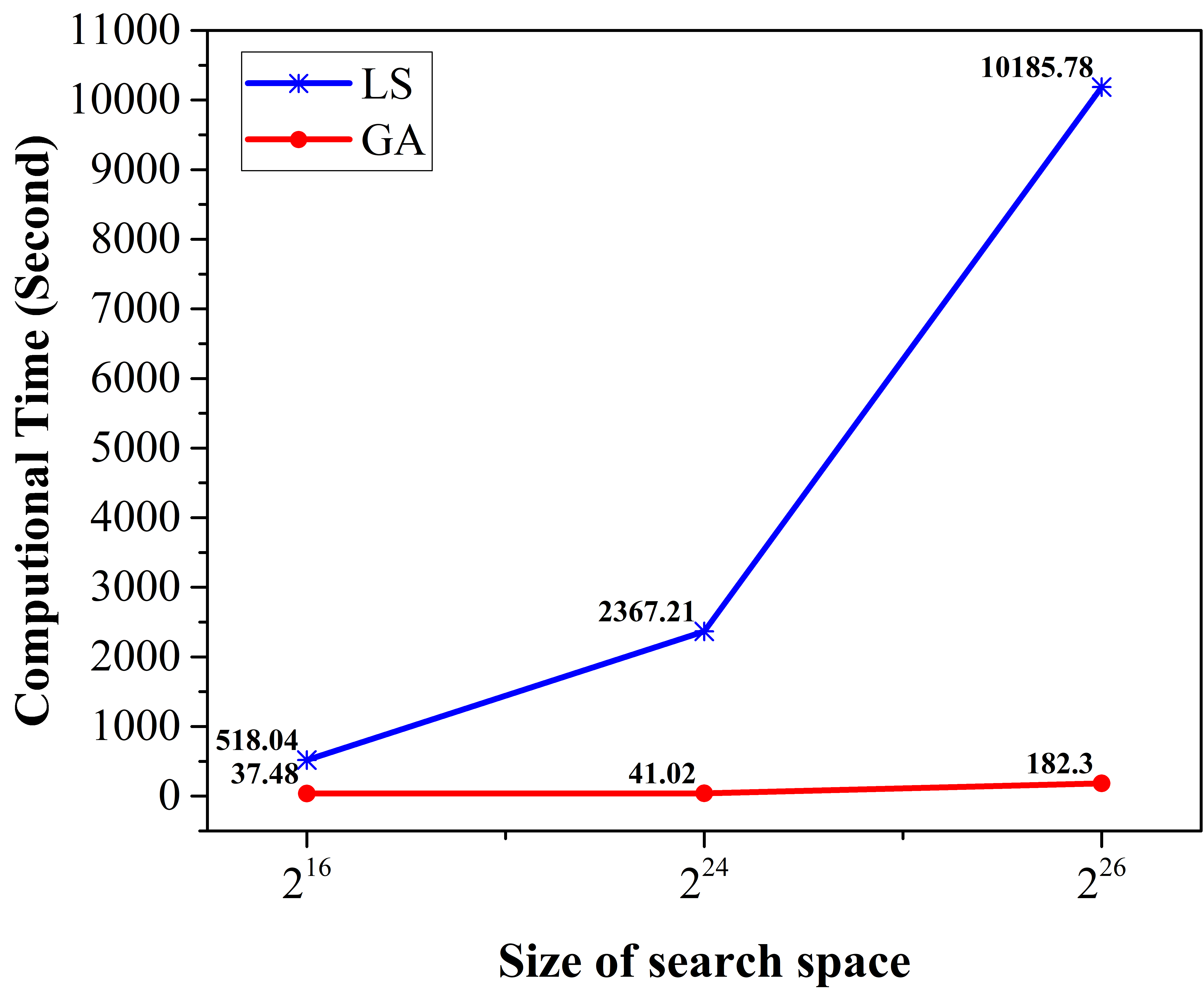}
\caption{Comparison of LS and GA \label{graph}}
\end{figure*}

Even though the linear search  is an effective method for finding solutions in the small-size search space, this is not the case for the big size search space, it is time consuming. On the other hand, optimization algorithms have received much attentions in recent years due to their success in many problems with large parameters.
Surprisingly, there has not been any attempt to use optimization methods for finding the bigger size self-dual codes until now.
In this section, it is shown that the heuristic approaches outperforms the linear search, by performing a search for extremal binary self-dual codes that have generator matrices of the $[I|\sigma(v)]$ form, using the two approaches. We use the same software - Magma \cite{Magma}, for the two approaches and  algorithms are run on a workstation with Intel Xeon 4.0 GHz processor and 64 GByte RAM. We compare the computational times and show that the GA finds codes significantly faster than the LS.
We consider the  sizes  $2^{16}$, $2^{24}$ and $2^{26}$ for the search field and try to find self-dual codes for the lengths 32, 48 and 52 using generator matrices of the $[I|\sigma(v)]$ form.
The experiments that we did for GA and LS show that the genetic algorithm outperforms the linear search. Even though, both algorithms find the same number of self-dual codes, GA has superiority in terms of the computational time.  One can directly see from Figure \ref{graph}  that when the search spaces grows, computational time increases suddenly in linear search. For GA, even if search space grows, the increase in the computational time in GA  is very small comparing to LS. Therefore GA is a time saving tool for the big size search space.

\newpage
\section{Computational Results}

In this section we employ the matrix defined in Equation~(\ref{MainMatrixConstruction}) and the GA algorithm to search for binary self-dual codes with parameters $[68,34,12]$ and $[72,36,12].$ That is, we search for these codes directly over the finite field $\mathbb{F}_2.$ We note that with the construction described in Section~2 and the alphabet $\mathbb{F}_2,$ the search field for codes of length 68 is $2^{34}$ and $2^{36}$ for codes of length 72. The searches are done in Magma \cite{Magma}. All the searches were completed in less than 24 hours. We split this section into subsections.

\subsection{Codes of length 68}

The weight
enumerator of a self-dual $[68,34,12]$ code is in one of the following
forms by \cite{AXVI, AXVII}:
\begin{equation*}
W_{68,1}=1+(442+4\beta)y^{12}+(10864-8\beta)y^{14}+\dots,
\end{equation*}
\begin{equation*}
W_{68,2}=1+(442+4\beta)y^{12}+(14960-8\beta-256\gamma)y^{14}+\dots \ ,
\end{equation*}
where $\beta$ and $\gamma$ are parameters and $0 \leq \gamma \leq 9.$ The following codes with the parameter $\gamma=0$ are known (\cite{Gulliver2, Yankov2, Korban1}):

$\gamma=0,\\
\beta \in \{2m|m=0,7,11,14,17,20,21,\dots
,99,100,102,105,110,119,136,165\};\ \text{or} \ \beta \in \{2m+1|m=3,5,8,10,15,16,17,19,20,\dots ,82,87,91,\dots,99,101,104,110,115,127\}$

\subsubsection{New singly even $[68,34,12]$ Codes}

\begin{enumerate}
\item[1.] The group $G=D_{34}$

Let $G=\langle x,y \ | \ x^{17}=y^2=1, x^y=x^{-1} \rangle\cong D_{34}.$ Let $v=\sum_{i=0}^{16} \sum_{j=0}^1 a_{1+i+17j}y^jx^i \in RD_{34},$ then
$$\sigma(v)=CIRC(A,B),$$
where $A=circ(a_1,a_2,a_3,\dots,a_{17}), B=rcirc(a_{18},a_{19},a_{20},\dots,a_{34})$ and $a_i \in R.$

We now employ a generator matrix of the form $[I \ | \ \sigma(v)],$ where $I$ is the $34 \times 34$ identity matrix, and $R=\mathbb{F}_2,$ to search for binary self-dual codes with parameters $[68,34,12].$ We only list codes that are new in the table below. Also, since $\sigma(v)$ is a matrix fully defined by the first row, we only list the first row of the matrix $A$ which we label as $r_A$ and the first row of the matrix $B$ which we label as $r_B.$

\begin{table}[h!]
\caption{New Type I $[68,34,12]$ Codes}
\resizebox{0.82\textwidth}{!}{\begin{minipage}{\textwidth}
\centering
\begin{tabular}{ccccccc}
\hline
         & Type       & $r_A$                                 & $r_B$                                 & $\gamma$ & $\beta$ & $|Aut(C_i)|$ \\ \hline
$C_{1}$ & $W_{68,2}$ & $(0,0,0,0,1,1,1,0,1,1,1,1,1,1,1,0,1)$ & $(1,1,0,0,1,1,1,1,0,0,1,1,1,0,0,1,0)$ & $0$      & $34$    & $34$         \\ \hline
$C_2$    & $W_{68,2}$ & $(1,1,0,0,1,1,1,1,0,0,1,0,0,0,1,1,1)$ & $(0,0,1,0,1,0,0,1,0,1,1,0,0,0,1,0,1)$ & $0$      & $51$    & $34$         \\ \hline
$C_3$    & $W_{68,2}$ & $(1,0,0,1,1,1,1,0,1,1,1,0,0,1,0,0,0)$ & $(1,0,1,0,1,1,1,1,0,1,1,0,1,1,1,0,1)$ & $0$      & $68$    & $34$         \\ \hline
$C_4$    & $W_{68,2}$ & $(0,0,1,0,0,0,0,0,0,1,0,1,1,0,0,0,0)$ & $(0,1,0,0,1,0,0,1,1,1,1,1,1,0,0,0,1)$ & $0$      & $85$    & $34$         \\ \hline
$C_5$    & $W_{68,2}$ & $(0,0,0,0,1,0,0,0,0,0,1,0,0,0,1,0,1)$ & $(0,0,1,0,1,1,1,1,0,1,0,1,0,0,0,1,1)$ & $0$      & $119$   & $34$         \\ \hline
$C_6$    & $W_{68,2}$ & $(0,0,0,1,0,1,0,1,1,1,1,1,0,1,0,1,1)$ & $(0,0,0,1,1,0,0,1,1,0,0,1,1,0,1,0,0)$ & $0$      & $153$   & $34$         \\ \hline
$C_7$    & $W_{68,2}$ & $(1,1,0,0,0,1,1,1,0,0,0,0,0,0,1,0,1)$ & $(1,1,0,1,0,1,0,0,0,0,0,0,0,0,1,1,0)$ & $0$      & $136$   & $34$         \\ \hline
$C_{8}$ & $W_{68,2}$ & $(0,1,1,0,0,0,0,1,1,0,0,0,1,0,1,0,1)$ & $(1,1,0,1,0,0,0,1,1,0,0,1,1,1,0,1,1)$ & $0$      & $187$   & $34$         \\ \hline
$C_{9}$ & $W_{68,2}$ & $(0,1,0,1,1,0,0,1,0,1,1,0,0,1,1,0,0)$ & $(1,1,1,0,1,0,1,1,1,1,0,0,0,0,0,1,0)$ & $0$      & $221$   & $34$         \\ \hline
$C_{10}$ & $W_{68,2}$ & $(0,0,0,0,1,1,0,0,0,0,0,1,1,1,0,0,0)$ & $(0,0,1,1,0,0,0,0,0,0,1,1,0,1,1,1,1)$ & $0$      & $238$   & $34$         \\ \hline
$C_{11}$ & $W_{68,2}$ & $(0,0,0,0,1,0,1,1,1,1,0,1,0,0,0,1,1)$ & $(1,0,1,0,1,1,1,0,1,0,1,0,0,0,0,1,1)$ & $0$      & $255$   & $34$         \\ \hline
\end{tabular}
\end{minipage}}
\end{table}

The codes with $\gamma=0$ and $\beta=34, 68, 136, 238$ were constructed in \cite{Gulliver2}, but there, the order of the automorphism group for all these codes is 68. In our case, these codes have the order of the automorphism group of 34 which indicates that they are new.

Similarly, codes with $\gamma=0$ and $\beta=51, 85, 119, 153, 187, 221, 255$ were constructed in \cite{Yankov2}, but all had a different order of the automorphism group to the codes we have obtained.

\end{enumerate}

\textbf{Note:} We have also tried searching for extremal binary self dual codes of length 68 using the generator matrix defined above with the linear search method, but the computations were still running after 3 days, so we have given up.

\subsection{Codes of length 72}

The possible weight enumerators for a Type I $[72,36,12]$ codes are as follows:
$$W_{72,1}=1+2\beta y^{12}+(8640-64\gamma)y^{14}+(124281-24\beta+384\gamma)y^{16}+\dots$$
$$W_{72,2}=1+2 \beta y^{12}+(7616-64 \gamma)y^{14}+(134521-24 \beta+384 \gamma)y^{16}+\dots$$
where $\beta$ and $\gamma$ are parameters. The following codes are known from \cite{Dougherty1, Kaya1, Dontcheva1, Gulliver1, Dougherty2, Yankov1, Yankov2, Yildiz1, Zhdanov1, Zhdanov2, Bouyukliev1}:

$\gamma=0,\\
\beta=87,96,114,141,147,153,159,162,171,177,180,183,185,189,195,196,199,200,201,203,204,\\
205,207,\dots,218,220,221,222,225,226,227,228,230,231,232,233,234,235,237,238,239,240,\\
242,\dots,247,249,251,252,254,255,257,258,259,261,262,264,265,267,271,273,275,276,279,\\
281,285,288,291,294,295,297,303,306,312,315,317,318,319,321,324,329,330,331,333,339,\\
341,348,351,353,355,357,360,363,365,375,377,379,381,387,389,393,396,402,403,405,413,\\
417,427,429,423,432,435,459,463,483,485,499,504,507,509,523,\dots,575,577,579,580,735,$

$\gamma=1,\\
\beta=193,195,199,200,206,207,\\
208,211,212,213,215,216,217,219,220,222,223,225,226,227,229,232,\dots,240,242,243,244,\\
246,247,248,249,250,252,254,256,257,258,260,261,264,266,270,274,276,277,525,526,527,\\
532,533,534,539,\dots,577,579,580,581,$

$\gamma=2,\\
\beta=195,199,201,218,219,222,223,228,231,\dots,233,239,240,241,243,244,245,250,251,255,\\
257,261,262,264,266,267,268,276,278,279,285,527,538,542,549,552,555,560,562,564,565,\\
566,568,\dots,573,575,576,580,584,585,$

$\gamma=3,\\
\beta=196,210,215,217,218,219,231,236,238,241,244,245,248,250,251,252,254,256,258,260,\\
261,262,266,267,268,270,272,273,276,280,284,294,297,548,552,558,562,568,581,582,$

$\gamma=4,\\
\beta=229,231,245,249,253,259,263,264,266,273,275,279,287,292,581,$

$\gamma=5,\\
\beta=228,229,231,234,235,236,242,249,255,259,265,266,269,273,\dots,278,281,283,285,286,\\
288,$

$\gamma=6,\\
\beta=145,153,159,165,169,171,177,181,183,193,195,201,207,213,217,219,225,231,237,243,\\
253,255,261,263,265,267,275,277,279,283,285,291,297,303,305,309,315,321,325,327,345,\\
349,351,387,411,423,$

$\gamma=7,\\
\beta=262,278,280,287,296,$

$\gamma=8,\\
\beta=277, 291,$

$\gamma=9,\\
\beta=267,287,307,311,$

$\gamma=10,\\
\beta=249,269,289,309,$

$\gamma=12,\\
\beta=319,359,379,$

$\gamma=16,\\
\beta=197,237,277,$

$\gamma=18,\\
\beta=176,186,206,236,246,276,296,$

$\gamma=24,\\
\beta=331,333,339,345,355,357,363,381,383,393,411,427,429,449,453,483,497,499,501,525,\\
573,$

$\gamma=36,\\
\beta=201,327,357,363,369,438,465,492,498,513,519,528,531,537,576,630,642,765,783,$

$\gamma=48,\\
\beta=629,653,$

in $W_{72,1},$ and the following ones:

$\gamma=0,\\
\beta \in \{1+11m \ | \ 7 \leq m \leq 48, \ m=51,52,54,56\},\\
 \beta \in \{11m \ | \ 6 \leq m \leq 44, \ m=47,59,68,74,76\}, \ \beta=88,89,102,119,132,136,153,154,165,170,\\
187,198,204,220,231,238,242,253,255,263,264,272,275,286,287,289,297,306,323,317,319,\\
330,335,340,357,363,374,385,391,418,425,442,459,462,476,483,493,527,$

$\gamma=11,\\
\beta \in \{1+11m \ | \ m=13 \leq m \leq 48, \ m=50,51,52,54,56,58,60,62,78\}, \beta \in \{11m \ | \ 13 \leq m \leq 54, \ m=56,58\},$

$\gamma=17,\\
\beta=238,255,272,289,306,323,340,357,374,391,408,425,442,459,476,493,510,527,544,595,\\
612,$

$\gamma=22,\\
\beta \in \{1+11m \ | \ 23 \leq m \leq 54\}, \beta \in \{11m \ | \ 23 \leq m \leq 57\},$

$\gamma=33,\\
\beta \in \{1+11m \ | \ 38 \leq m \leq 57, \ m=59,60,61,62\}, \beta \in \{11m \ | \ 36 \leq m \leq 38, 40 \leq m \leq 58,60 \leq m \leq 64\},$

$\gamma=34,\\
\beta=306,340,391,408,425,442,459,476,493,510,527,544,561,578,612,646,663,680,697,714,$

$\gamma=44,\\
\beta \in \{606,617,727\}, \beta \in \{605,627,737\},$

$\gamma=71,\\
\beta=765,$

in $W_{72,2}.$ The possible weight enumerators for Type II $[72,36,12]$ codes are \cite{Dougherty1}
$$1+(4398+\alpha)y^{12}+(197073-12\alpha)y^{16}+(18396972+66\alpha)y^{20}+\dots$$
The following codes are known from \cite{Dougherty1, Kaya1, Kaya2, Dontcheva1, Dontcheva2, Gulliver1, Dougherty2, Yankov1, Yankov2, Yildiz1, Zhdanov1, Zhdanov2, Bouyukliev1}:

$\alpha=-3744, -3774, -3768, -3714, -3762, -3792, -3732, -3702, -3756,-3750, -3738, -3726,\\
-3708, -3720, -3786, -1416, -3810, -3798, -3828,-3678, -3816, -3846, -3654, -3648, -3690,\\
 -3822, -3696, -3660, -3684, -3642, -3672, -3936, -1356, -3600,-1362, -3846, -3570, -3708,\\
 -3294, -3432, -3156, -3018, -3984,-4008, -3978, -3972, -3936, -3924, -3894,
-3870, -3846,\\
 -3840, -3834, -3828, -3816, -3810, -3798, -3786, -3768, -3756, -3750, -3744, -3732, -3726,\\ -3720, -3714, -3708, -3696, -3684,-3678, -3672, -3666, -3660, -3648, -3642, -3636, -3630,\\
-3624, -3612,
-3606, -3600, -3594 ,-3588, -3582, -3576, -3570, -3564, -3558, -3552,-3546,\\
-3540, -3534,-3528, -3522, -3516, -3510, -3504, -3498, -3492,-3486, -3474, -3468, -3462,\\
-3456, -3450, -3444, -3438, -3432, -3426,-3420, -3414, -3408, -3402, -3384, -3378, -3372,\\
-3366, -3360, -3354,-3348, -3342, -3336, -3330, -3324, -3318, -3312, -3306, -3300, -3294,\\
-3288, -3282, -3276, -3270, -3264, -3258, -3252, -3246, -3240, -3234,-3228, -3222, -3216,\\
-3210, -3204, -3198, -3192, -3186, -3180, -3174,-3168, -3162, -3144, -3138, -3126, -3120,\\
-3114, -3108, -3102, -3096,-3090, -3084, -3078, -3072, -3054, -3042, -3036, -3018, -3012,\\
-3006,-3000, -2994, -2982, -2976, -2970, -2958, -2952, -2946, -2934, -2928,
-2922, -2916,\\
-2910, -2904, -2898, -2892, -2886, -2880, -2874, -2856, -2838, -2826, -2820, -2814, -2771,\\
-2742, -2730, -2724, -2718, -2694, -2676, -2670, -2652, -2646, -2640, -2616, -2592, -2586,\\
-2568, -2502, -2466, -2394, -2340, -1416, -3600, -3576, -3552, -3546, -3540, -3534, -3528,\\
-3522, -3510, -3504, -3498, -3492, -3480, -3468, -3462, -3456, -3444, -3432, -3420, -3408,\\
-3396, -3384, -3372, -3348, -3336, -3300, -3228, -3204, -2316, -4134, -4068, -4062, -4002,\\
-3996, -3930, -3804, -2940, -2808, -2748, -2610, -2544, -2418, -2412, -2016, -4092, -3990,\\
-3888, -2868, -2766, -2664, -2562, -3996, -3900, -3888, -3876, -3852, -3804, -3768, -3756,\\
-3744, -3732, -3708, -3696, -3672, -3660, -3624, -3612, -3600, -3588, -3564, -3552, -3516,\\
-3492, -3480, -3468, -3456, -3444, -3384, -3336, -3312, -3300, -3264, -3252, -3192, -3180,\\
-3156, -3120, -3036, -3024, -2976, -2952, -2868, -4134, -4002, -3996, -3870, -3864, -3804,\\
-3738, -3732, -4672, -3606, -3600, -3540, -3474, -3468, -3408, -3342, -3336, -3276, -3210,\\
-3204, -3144, -3078, -3072, -2946, -2940, -2808, -3960, -3840, -3732, -3720, -3612, -3600,\\
-3492, -3480, -3372, -3360, -3252, -3240, -3120, -3000, -2880, -3942, -3882, -3822, -3786,\\
-3762, -3702, -3642, -3606, -3582, -3546, -3522, -3462, -3402, -3366, -3306, -3246, -3186,\\
-2424, -2784, -2712, -2580, -2832, -2532, -2964, -3060, -2988, -2736, -3048, -2844, -3132,\\
-2820, -2748, -3096, -3426, -3480, -3390, -3354, -3444, -3408, -3372, -3336, -3378, -3432,\\
-3468, -3558, -3414, -3276, -3312, -3258, -3198, -3450, -3360, -3324, -3021, -3252, -3294,\\
-3300, -3384, -3264, -3462, -3228, -3456, -3420, -3402, -3474, -3282, -3552, -3366, -3330,\\
-3204, -3318, -3348, -3534, -3486, -3522, -3270, -3492, -3492, -3396, -3222, -3240, -3078,\\
-3438, -3546, -3498, -3306, -3510, -3180, -3540, -3504, -3192, -3516, -3162, -3234, -3072,\\
-3684, -3090, -2910, -2928, -3564, -3210, -3156, -2736, -2748, -2844, -2964, -3060, -3396,\\
-2682, -2700, -2754, -2790, -2802, -2862, -2988, -3132, -3150, -3654, -3690, -3774, -3780,\\
-3792, -3906, -3918, -4069, -4086, -3618.$

\subsubsection{New singly even and doubly even self-dual $[72,36,12]$ codes}

\begin{enumerate}
\item[1.] The group $G=C_2 \times C_{18}$

Let $G=\langle x,y \ | \ x^{18}=y^2=1, xy=yx \rangle \cong C_2 \times C_{18}.$ Let $v=\sum_{i=0}^{17} \sum_{j=0}^1 a_{1+i+18j} x^iy^j \in R(C_2 \times C_{18}),$ then
$$\sigma(v)=CIRC(A,B),$$
where $A=circ(a_1,a_2,a_3,\dots,a_{18}), B=circ(a_{19},a_{20},a_{21},\dots,a_{36})$ and $a_i \in R.$

We now employ a generator matrix of the form $[I \ | \ \sigma(v)],$ where $I$ is the $36 \times 36$ identity matrix, and $R=\mathbb{F}_2,$ to search for binary self-dual codes with parameters $[72,36,12].$ We only list codes that are new in the table below. Also, since $\sigma(v)$ is a matrix fully defined by the first row, we only list the first row of the matrix $A$ which we label as $r_A$ and the first row of the matrix $B$ which we label as $r_B.$

\begin{table}[htbp]
\caption{New Type I $[72,36,12]$ Codes}
\resizebox{0.8\textwidth}{!}{\begin{minipage}{\textwidth}
\centering
\begin{tabular}{ccccccc}
\hline
      & Type       & $r_A$                                   & $r_B$                                   & $\gamma$ & $\beta$ & $|Aut(C_i)|$ \\ \hline
$C_1$ & $W_{72,1}$ & $(1,1,1,1,0,0,1,1,1,1,0,0,0,0,1,0,1,0)$ & $(0,0,0,1,0,0,1,0,0,1,1,0,1,0,1,1,0,0)$ & $0$      & $201$   & $72$         \\ \hline
$C_2$ & $W_{72,1}$ & $(1,1,1,0,0,1,1,1,1,1,1,0,1,0,0,1,0,1)$ & $(1,0,1,1,0,1,1,1,1,0,0,1,0,0,1,0,0,0)$ & $36$     & $471$   & $72$         \\ \hline
\end{tabular}
\end{minipage}}
\end{table}
The code $C_1$ has parameters $\gamma=0, \beta=201$ - a code with these parameters was constructed in \cite{Yildiz1} but there, the order of the automorphism group was $48$, in our case the order of the automorphism group is $72$ which indicates that the code is new.

\item[2.] The group $G=C_{18,2}$

Let $G=\langle x \ | \ x^{18 \cdot 2}=1 \rangle \cong C_{18,2}.$ Let $v=\sum_{i=0}^{17} \sum_{j=0}^{1} a_{1+i+18j}x^{2i+j} \in R(C_{18,2}),$ then
$$\sigma(v)=\begin{pmatrix}
A&B\\
A'&B
\end{pmatrix},$$
where $A=circ(a_1,a_2,a_3,\dots,a_{18}), B=circ(a_{19},a_{20},a_{21},\dots,a_{36}), A'=circ(a_{36},a_{19},\\
a_{20},\dots,a_{35})$ and $a_i \in R.$

We now employ a generator matrix of the form $[I \ | \ \sigma(v)],$ where $I$ is the $36 \times 36$ identity matrix, and $R=\mathbb{F}_2,$ to search for binary self-dual codes with parameters $[72,36,12].$ We only list codes that are new in the table below. Also, since $\sigma(v)$ is a matrix fully defined by the first row, we only list the first row of the matrix $A$ which we label as $r_A$ and the first row of the matrix $B$ which we label as $r_B.$

\begin{table}[htbp]
\caption{New Type I $[72,36,12]$ Codes}
\resizebox{0.8\textwidth}{!}{\begin{minipage}{\textwidth}
\centering
\begin{tabular}{ccccccc}
\hline
      & Type       & $r_A$                                   & $r_B$                                   & $\gamma$ & $\beta$ & $|Aut(C_i)|$ \\ \hline
$C_3$ & $W_{72,1}$ & $(1,0,0,0,0,0,1,1,0,1,1,1,1,0,1,1,0,0)$ & $(1,0,0,1,0,0,0,1,1,1,1,1,1,1,1,0,1,1)$ & $72$      & $825$   & $72$         \\ \hline
\end{tabular}
\end{minipage}}
\end{table}

We note that a Type I $[72,36,12]$ code with $\gamma=72$ in $W_{72,1}$ was not constructed before.

\item[3.] The group $G=C_4 \times C_9$

Let $G=\langle x,y \ | \ x^{9}=y^4=1, xy=yx \rangle \cong C_4 \times C_{9}.$ Let $v=\sum_{i=0}^{8} \sum_{j=0}^3 a_{1+i+9j} x^iy^j \in R(C_4 \times C_{9}),$ then
$$\sigma(v)=CIRC(A,B,C,D),$$
where $A=circ(a_1,a_2,a_3,\dots,a_{9}), B=circ(a_{10},a_{11},a_{12},\dots,a_{18}), C=circ(a_{19},a_{20},\\
a_{21},\dots,a_{27}), D=circ(a_{28},a_{29},a_{30},\dots,a_{36})$ and $a_i \in R.$

We now employ a generator matrix of the form $[I \ | \ \sigma(v)],$ where $I$ is the $36 \times 36$ identity matrix, and $R=\mathbb{F}_2,$ to search for binary self-dual codes with parameters $[72,36,12].$ We only list codes that are new in the tables below. Also, since $\sigma(v)$ is a matrix fully defined by the first row, we only list the first row of the matrices $A, B, C, D$ which we label as $r_A, r_B, r_C$ and $r_D$ respectively.

\begin{table}[htbp]
\caption{New Type I $[72,36,12]$ Codes}
\resizebox{0.75\textwidth}{!}{\begin{minipage}{\textwidth}
\centering
\begin{tabular}{ccccccccc}
\hline
      & Type       & $r_A$                 & $r_B$                 & $r_C$                 & $r_D$                 & $\gamma$ & $\beta$ & $|Aut(C_i)|$ \\ \hline
$C_4$ & $W_{72,1}$ & $(0,1,0,0,1,1,0,1,0)$ & $(1,0,0,1,1,1,1,0,0)$ & $(0,0,0,1,0,1,1,1,0)$ & $(0,1,1,1,0,0,1,0,0)$ & $36$     & $441$   & $72$         \\ \hline
\end{tabular}
\end{minipage}}
\end{table}

\begin{table}[h!]
\caption{New Type II $[72,36,12]$ Codes}
\resizebox{0.82\textwidth}{!}{\begin{minipage}{\textwidth}
\centering
\begin{tabular}{ccccccc}
\hline
      & $r_A$                 & $r_B$                 & $r_C$                 & $r_D$                 & $\alpha$ & $|Aut(C_i)|$ \\ \hline
$C_5$ & $(0,0,0,1,0,1,1,1,0)$ & $(1,0,0,1,1,0,1,1,0)$ & $(1,1,1,1,0,0,1,0,0)$ & $(1,0,1,0,1,1,0,1,0)$ & $-2772$  & $72$         \\ \hline
\end{tabular}
\end{minipage}}
\end{table}

\item[4.] The group $G=C_3 \times C_{12}$

Let $G=\langle x,y \ | \ x^{12}=y^3=1, xy=yx \rangle \cong C_3 \times C_{12}.$ Let $v=\sum_{i=0}^{11} \sum_{j=0}^2 a_{1+i+12j} x^iy^j \in R(C_3 \times C_{12}),$ then
$$\sigma(v)=CIRC(A,B,C),$$
where $A=circ(a_1,a_2,a_3,\dots,a_{12}), B=circ(a_{13},a_{14},a_{15},\dots,a_{24}), C=circ(a_{25},a_{26},\\
a_{27},\dots,a_{36})$ and $a_i \in R.$

We now employ a generator matrix of the form $[I \ | \ \sigma(v)],$ where $I$ is the $36 \times 36$ identity matrix, and $R=\mathbb{F}_2,$ to search for binary self-dual codes with parameters $[72,36,12].$ We only list codes that are new in the tables below. Also, since $\sigma(v)$ is a matrix fully defined by the first row, we only list the first row of the matrices $A, B, C$ which we label as $r_A, r_B$ and $r_C$ respectively.

\begin{table}[h!]
\caption{New Type I $[72,36,12]$ Codes}
\resizebox{0.78\textwidth}{!}{\begin{minipage}{\textwidth}
\centering
\begin{tabular}{cccccccc}
\hline
         & Type       & $r_A$                       & $r_B$                       & $r_C$                       & $\gamma$ & $\beta$ & $|Aut(C_i)|$ \\ \hline
$C_{6}$ & $W_{72,1}$ & $(1,0,0,0,0,1,1,1,1,1,1,0)$ & $(1,0,0,1,1,0,0,1,0,1,1,0)$ & $(0,0,1,0,1,1,1,1,1,0,1,1)$ & $36$     & $456$   & $72$         \\ \hline
\end{tabular}
\end{minipage}}
\end{table}

\begin{table}[h!]
\caption{New Type II $[72,36,12]$ Codes}
\resizebox{0.835\textwidth}{!}{\begin{minipage}{\textwidth}
\centering
\begin{tabular}{cccccc}
\hline
         & $r_A$                       & $r_B$                       & $r_C$                       & $\alpha$ & $|Aut(C_i)|$ \\ \hline
$C_7$    & $(1,1,0,1,0,0,0,1,1,1,1,1)$ & $(1,1,1,1,1,1,0,1,0,0,0,1)$ & $(1,0,1,0,1,1,1,0,1,0,1,0)$ & $-2106$  & $144$        \\ \hline
$C_8$    & $(1,0,1,0,1,0,0,1,1,1,1,0)$ & $(0,0,0,1,1,1,1,1,1,1,1,0)$ & $(0,1,1,1,1,0,1,1,1,1,0,0)$ & $-2322$  & $144$        \\ \hline
$C_9$    & $(0,1,1,1,0,0,1,1,1,1,1,1)$ & $(0,0,0,1,0,0,0,0,0,0,0,1)$ & $(0,1,0,0,1,1,0,0,1,0,0,0)$ & $-2472$  & $144$        \\ \hline
$C_{10}$    & $(0,1,1,0,1,0,1,1,0,1,0,0)$ & $(0,1,1,1,1,0,0,0,0,1,1,0)$ & $(0,1,1,1,1,1,0,1,0,1,0,0)$ & $-2520$  & $144$        \\ \hline
$C_{11}$ & $(0,1,0,1,1,0,0,1,1,1,0,0)$ & $(0,1,1,1,1,1,1,1,1,0,0,0)$ & $(0,1,0,1,0,0,1,1,1,0,0,0)$ & $-2550$  & $144$        \\ \hline
$C_{12}$ & $(1,0,1,1,0,0,0,0,1,1,0,0)$ & $(1,0,1,1,0,1,1,1,0,0,0,0)$ & $(0,0,0,1,0,1,0,1,1,0,0,0)$ & $-3030$  & $72$         \\ \hline
$C_{13}$ & $(0,1,0,0,1,1,0,1,0,1,0,0)$ & $(0,0,1,1,1,0,1,0,1,0,1,0)$ & $(1,1,1,1,1,0,0,1,0,1,0,1)$ & $-3954$  & $144$        \\ \hline
\end{tabular}
\end{minipage}}
\end{table}

\item[5.] The group $G=D_{36}$

Let $G=\langle x,y \ | \ x^{18}=y^2=1, x^y=x^{-1} \rangle\cong D_{36}.$ Let $v=\sum_{i=0}^{17} \sum_{j=0}^1 a_{1+i+18j}y^jx^i \in RD_{36},$ then
$$\sigma(v)=CIRC(A,B),$$
where $A=circ(a_1,a_2,a_3,\dots,a_{18}), B=rcirc(a_{19},a_{20},a_{21},\dots,a_{36})$ and $a_i \in R.$

We now employ a generator matrix of the form $[I \ | \ \sigma(v)],$ where $I$ is the $36 \times 36$ identity matrix, and $R=\mathbb{F}_2,$ to search for binary self-dual codes with parameters $[72,36,12].$ We only list codes that are new in the table below. Also, since $\sigma(v)$ is a matrix fully defined by the first row, we only list the first row of the matrix $A$ which we label as $r_A$ and the first row of the matrix $B$ which we label as $r_B.$

\begin{table}[h!]
\caption{New Type I $[72,36,12]$ Codes}
\resizebox{0.8\textwidth}{!}{\begin{minipage}{\textwidth}
\centering
\begin{tabular}{ccccccc}
\hline
      & Type       & $r_A$                                   & $r_B$                                   & $\gamma$ & $\beta$ & $|Aut(C_i)|$ \\ \hline
$C_{14}$ & $W_{72,1}$ & $(1,1,0,0,1,1,1,1,0,1,1,0,1,0,0,1,1,1)$ & $(0,1,0,1,0,1,1,0,1,1,1,0,0,0,0,0,1,1)$ & $0$      & $354$   & $36$         \\ \hline
$C_{15}$ & $W_{72,1}$ & $(1,0,0,1,1,0,0,1,1,1,1,1,0,1,0,1,0,0)$ & $(0,1,1,1,0,1,1,0,1,0,1,0,1,0,0,1,1,1)$ & $18$      & $273$   & $36$         \\ \hline
$C_{16}$ & $W_{72,1}$ & $(0,0,0,0,0,1,0,0,1,1,1,1,0,0,0,1,0,0)$ & $(0,0,1,1,0,1,1,0,1,1,1,1,1,0,1,1,0,0)$ & $36$      & $372$   & $36$         \\ \hline
$C_{17}$ & $W_{72,1}$ & $(0,1,0,0,1,0,0,1,1,0,1,0,0,0,1,1,1,1)$ & $(1,0,1,1,0,1,0,0,0,0,1,0,1,0,0,1,0,1)$ & $54$      & $669$   & $36$         \\ \hline
\end{tabular}
\end{minipage}}
\end{table}

We note that a Type I $[72,36,12]$ code with $\gamma=54$ in $W_{72,1}$ was not constructed before.

\end{enumerate}

\newpage

\textbf{Note:} Similarly as for codes of length 68 in the previous section, we have also tried searching for binary self dual codes of length 72 using the generator matrices defined above with the linear search method, but the computations were still running after 7 days, so we have given up.

\section{Conclusion}

In this work, we explored the genetic algorithm in the search of binary self-dual codes. We employed a matrix construction from \cite{Gildea1}, and searched for binary self-dual codes of different lengths using a standard linear search and the genetic algorithm. We showed that the latter, copes with large search fields significantly better and finds codes much faster then the standard linear search. Moreover, with the matrix construction from \cite{Gildea1} and the genetic algorithm, we found new binary self-dual codes. In particular, we found 11 extremal binary self-dual codes of length 68 and 17 new binary self-dual codes of length 72 with the minimum distance 12. We stress that the searches were done in a very short time and directly over the finite field $\mathbb{F}_2,$ that is, we have not considered any extension or neighbour methods as some authors have done in the literature - this highlights the effectiveness of the genetic algorithm. A suggestion for future work is to consider different matrix constructions together with the genetic algorithm to search for extremal binary self-dual codes of different lengths. Another possible direction is to consider the genetic algorithm with different alphabets than the finite field $\mathbb{F}_2$ when searching for binary self-dual codes.

\end{document}